\documentclass[runningheads]{llncs}
\usepackage{url}
\usepackage[T1]{fontenc}
\usepackage[sort,compress]{cite} 
\usepackage[colorlinks=true, urlcolor=blue, linkcolor=blue, citecolor=blue]{hyperref}
\usepackage{graphicx}
\begin{document}
\title{The Space Between Us: A Methodological Framework for Researching Bonding and Proxemics in Situated Group-Agent Interactions}
\titlerunning{The Space Between Us}
%
\author{Ana Müller\orcidID{0000-0002-4960-082X} \and
Anja Richert\orcidID{0000-0002-3940-3136}}
\authorrunning{Müller and Richert}
%
\institute{Cologne Cobots Lab, TH Köln — University of Applied Sciences, Cologne, Germany \\
\email{ana.mueller@th-koeln.de}, \url{www.th-koeln.de}}
\maketitle              

\begin{abstract}
This paper introduces a multimethod framework for studying spatial and social dynamics in real-world group-agent interactions with socially interactive agents. Drawing on proxemics and bonding theories, the method combines subjective self-reports and objective spatial tracking. Applied in two field studies in a museum (\emph{N} = 187) with a robot and a virtual agent, the paper addresses the challenges in aligning human perception and behavior. We focus on presenting an open source, scalable, and field-tested toolkit for future studies.
\keywords{group interactions \and multiparty \and methodology \and in-the-wild}
\end{abstract}

\section{Motivation}

\begin{figure}[htbp!]
    \centering
    \includegraphics[width=1\linewidth]{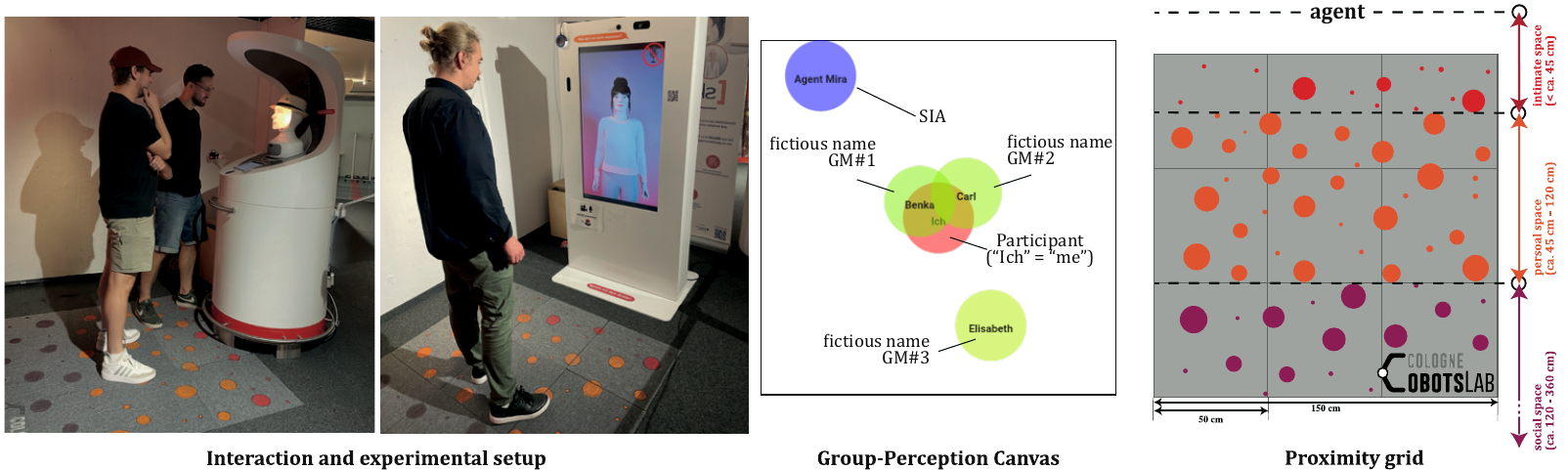}
    \caption{Left: Interactions with agents at the museum in 2024. (i) With Furhat robot (Furhat Robotics Sweden); (ii) with virtual agent (MetaHuman); (iii) Example result from the Group-Perception Canvas; (iv) Setup of the proximity grid.}
    \label{figinteractions}
\end{figure}

Social life unfolds in groups, shaping identity, behavior, and interaction in shared spaces \cite{levineSmallGroups2006}. Previous studies on human-agent interaction (HAI) have focused mainly on dyadic contexts under controlled conditions, overlooking the intricacies of group-agent interactions (GAI) in in-the-wild settings \cite{frauneHumanGroupPresence2019,seboRobotsGroupsTeams2020,oliveiraHumanRobotInteractionGroups2021,gilletMultipartyInteractionHumans2022,jungRobotsWildTime2018}. However, with the growing use of socially interactive agents in (semi-)public spaces, they engage with groups rather than individuals \cite{sabanovicWereThisTogether2020,mullerNoOneIsland2023,oliveiraHumanRobotInteractionGroups2021}, and dyadic models quickly become insufficient \cite{jungRobotsWildTime2018,abramsICEFrameworkConcepts2020}. GAIs introduce additional complexity: beyond individual dynamics, group composition and intragroup relationships influence how agents are perceived and engaged with \cite{gilletInteractionShapingRoboticsRobots2024,seboRobotsGroupsTeams2020,gilletMultipartyInteractionHumans2022}. This raises questions about how social constructs, such as proximity and bonding, are manifested in situated interactions. Although these constructs are central to GAI, robust methods for capturing them in-the-wild remain scarce \cite{oliveiraHumanRobotInteractionGroups2021,jungRobotsWildTime2018}. This paper addresses this gap by introducing a multimethod approach to assess proximity and bonding. The methods were applied and refined in two field studies at the Deutsches Museum Bonn (Germany) with a robot and a virtual agent (Fig. \ref{figinteractions}; i,ii). We outline the foundational reasoning, technical hurdles, and methodological refinements. While empirical findings are not the primary focus here, the framework offers a scalable and ecologically valid open-source methodological toolkit.

\section{Theoretical Background}

The research \& development of socially interactive agents increasingly requires attention to social contexts. Recent studies underscore the impact of group dynamics, including identity, cohesion, and peer presence, on perceptions \cite{gilletInteractionShapingRoboticsRobots2024,seboRobotsGroupsTeams2020,frauneHumanGroupPresence2019,gilletMultipartyInteractionHumans2022}. To date, controlled laboratory studies and Wizard-of-Oz setups \cite{riekWizardOzStudies2012} provide important knowledge; however, their ecological validity is limited. Deployments in real world settings add spontaneity, variability, and social interdependence \cite{jungRobotsWildTime2018,oliveiraHumanRobotInteractionGroups2021,mullerNoOneIsland2023}. Dyadic models, although foundational, fail to capture the dynamic aspects of GAI \cite{gilletMultipartyInteractionHumans2022}. In addition, techniques for evaluating essential social constructs, such as proximity and bonding, remain underdeveloped in the in-the-wild \cite{jungRobotsWildTime2018,oliveiraHumanRobotInteractionGroups2021}.

\textbf{Bonding} refers to the emotional or social relationship between a person and an agent. Related constructs include rapport \cite{lin2025connectioncoordinationrapportccrscale} and attitudes toward agents (e.g., \cite{ninomiyaDevelopmentMultidimensionalRobot2015}), often measured retrospectively in surveys \cite{oliveiraHumanRobotInteractionGroups2021}. However, most instruments were designed for dyadic interactions and do not account for GAI \cite{oliveiraHumanRobotInteractionGroups2021}. An exception is the Interactive IOS (Inclusion of Other in the Self) Scale for Multiparty Interactions (IIMI) \cite{seboInteractiveIOSScale2024,zhangIceBreakingTechnologyRobots2023}. Although promising, such tools have only been used in controlled research settings.

\textbf{Proxemics.} Hall's (1966) proxemics theory identifies four interpersonal zones in human-human interactions: intimate, personal, social, and public \cite{hallHiddenDimension1966}. These are unconsciously managed and influenced, e.g., by familiarity and context \cite{pattersonSpatialFactorsSocial1968}. Such frameworks have guided the development of socially aware robots \cite{pathiDetectingGroupsEstimating2022,mavrogiannisCoreChallengesSocial2023,moujahidComeCloserEffects2023,moujahidDemonstrationRobotReceptionist2022}, emphasizing the importance of adaptable tools for spatial behavior research.

\textbf{Linking Bonding and Proxemics.} Although conceptually distinct, concepts can be related: people approach agents they feel close to or conversely maintain distance from those they do not. However, most prior research on HAI/GAI has focused on objective spatial behavior or subjective closeness, without linking the two. To address this, we propose a quantitative multimethod approach that combines behavioral (i.e., proximity) and survey (i.e., bonding) data.

\section{Method Development Description}

We developed this framework to understand bonding, proximity, and their correlations in GAI. It was refined iteratively in two field studies (\emph{N} = 187). Of the total of 187 recorded interactions, approximately 55\% were GAI, comprising dyads (i.e., two humans), triads, and groups of four people. The method used a multimodal data collection approach: spatial video analysis was used to track proximity, while bonding was measured using an adaptive survey customized to the composition of the group. The participants independently engaged with the agent located at the entrance to the exhibition in an accessible semi-enclosed area (Fig. \ref{figinteractions}; i, ii). According to GDPR, agents were inactive until participants pressed a buzzer to provide their informed consent to data processing and interaction recording. The project was reviewed and approved by the Ethics Research
Committee of TH Köln (THK-2023-0004).

\textbf{Bonding.} After independently interacting with the agents, a researcher invited the users to participate in a survey on a tablet using SoSci Survey (Germany). The survey collected e.g., demographic information, previous experience and attitudes towards agents, along with measures of perceived group dynamics, interaction quality, and social aspects using a modified version of the Group Attitude Scale \cite{evansGroupAttitudeScale1986}(\href{https://github.com/anammueller/questionnaire-transcripst-Skilled/blob/main/Transkript%20des%20Fragebogens%20zur%20Studie%20im%20Deutschen%20Museum%20Bonn%20mit%20dem%20MetaHuman.pdf}{transcript on GitHub}). To assess bonding, we developed the \href{https://github.com/anammueller/group-perception-canvas}{Group Perception Canvas (GPC)}. At the start of the survey, participants were asked how many other people were present during the interaction. Later, this was shown back to them for confirmation (e.g., “You indicated that 3 other people were present. Is that correct?”). If necessary, they could revise their response. Afterwards they were asked to assign fictional names to each group member (GM), providing an additional opportunity to confirm or adjust the group size. These inputs directly informed the visualization logic: the number of circles representing GMs in the perceived bonding-task matched the final input of the participant. Participants used a drag-and-drop interface to arrange circles representing themselves, the agent, and GMs on a 500×500 px canvas (Fig. \ref{figinteractions}; iii). The task adapted dynamically to the reported group size and prompted participants with: “How close did you feel to the agent and the other participants? Position the circles — the closer and the greater the overlap, the closer you felt.” This resulted in an intuitive visual representation of perceived social closeness. The distances between the circles were calculated post hoc and validated through sample-based verification, which showed an average error of 1.36 mm (\emph{SD} = .80). The resulting metrics were included in the survey dataset. In parallel to survey participation, researchers completed an observation protocol documenting group size and visual identifiers (e.g., clothing) to allow accurate mapping between survey (i.e., bonding) and video (i.e., proximity) data and to establish a ground truth on the group size. 

\textbf{Proximity} was assessed using a spatial coding system applied to video recordings of interactions. A 150×150 cm proximity grid - based on Hall \cite{hallHiddenDimension1966} - was embedded in the physical setup using color-coded floor markers (Fig. \ref{figinteractions}; i,ii,iv). These visual elements were integrated to avoid priming participants (e.g., suggesting where to stand for interactions). Therefore, the inclusion of the institution logo helped the grid blend in with the environment as part of the exhibition design. As manual coding with qualitative software (e.g., MAXQDA) was considered impractical for this research design, and automated approaches such as LiDAR sensors do not allow one to distinguish between individual group members, the recordings were quantitatively annotated using \href{https://github.com/anammueller/Group-Proximity-Annotation-Tool-for-Human-Agent-Interaction}{Group-Proximity-Annotation-Tool (GPAT)} - a Python-based pipeline that combines OpenCV and Pandas. The interface allowed to preview clips and take notes (e.g., based on an observation protocol to distinguish participants from others). In every four frames, the coders annotated the location of each person using predefined labels corresponding to the colors in the proximity grid: intimate [red] (i), personal [orange] (p), social [purple] (s), or off-screen (x). Consistency was found to be high in the intracoder reliability checks conducted over several sessions. The metrics considered included the percentage of time spent within each zone, the predominant zone, and transitions for each member. The latter were algorithmically smoothed, such as by removing ‘x’ as an initial value, since it reflected a default state rather than meaningful user behavior. Afterwards, the variables were integrated into the survey dataset.

\textbf{Data Triangulation and Integration.} The variables of bonding (subjective data) and proximity (objective behavior) were combined in the survey dataset to explore their interrelation. Since proximity was measured as the time spent in each zone (in \%), while bonding was captured through continuous metrics (in mm), the variables were reaggregated and z-standardized to allow comparison \cite{fieldDiscoveringStatisticsUsing2024}. Initial analyses (e.g., correlations) suggested that these dimensions capture complementary aspects of interaction, underscoring the value of methodological triangulation. Integrating additional factors such as agent type, duration of interaction, and other variables can offer nuanced insights into GAI.

\section{Conclusion and Future Directions}
This paper presents a multimethod framework that integrates spatial tracking and adaptive self-report tools to examine proximity and bonding in real-world GAI. Although developed for GAI, the framework is flexible: it supports dyadic and multiparty interactions, can be used in both laboratory and field settings, and accommodates groups of varying sizes. The tools developed - the \href{https://github.com/anammueller/group-perception-canvas}{Group Perception Canvas (GPC)} and the \href{https://github.com/anammueller/Group-Proximity-Annotation-Tool-for-Human-Agent-Interaction}{Group-Proximity-Annotation-Tool (GPAT)} – are openly available and can be used independently or in combination, depending on the research context. A particularly promising use case involves collecting bonding reports from multiple group members, allowing researchers to compare subjective perceptions and relate them to corresponding behavioral data — offering novel insights into intragroup dynamics. However, the framework also faces limitations. Linking survey and video data currently requires extensive manual effort, and the \href{https://github.com/anammueller/Group-Proximity-Annotation-Tool-for-Human-Agent-Interaction}{GPAT} lacks real-time correction features, which complicates the coding process. Future work will address these challenges by automating data integration and enhancing annotation functionality. Future adaptations of the framework could also include mobile robots, which introduce additional challenges such as changing spatial reference frames. Ongoing updates to the tools and documentation will be made available through the public repositories, supporting future research on real-world GAI. 

\begin{credits}
\subsubsection{\ackname} This research was funded by the Federal Ministry of Research, Technology and Space, Germany, in the framework FH-Kooperativ 2-2019 (project number 13FH504KX9). We thank our collaboration partners DB Systel GmbH, Deutsches Museum Bonn and all other collaborators for their assistance and contributions.

\subsubsection{\discintname}
The authors have no competing interests to declare that are relevant to the content of this article.
\end{credits}

\bibliographystyle{IEEEtran}
\bibliography{Bib}
%
%
%
\end{document}